\documentclass[conference]{IEEEtran}

\usepackage[a4paper, total={7.3in, 11in}]{geometry}
\IEEEoverridecommandlockouts

\usepackage{cite}
\usepackage{algorithmicx}
\usepackage{amsmath,amssymb,amsfonts}
\usepackage{microtype}
\usepackage{graphicx}
\usepackage{subfigure}
\usepackage{booktabs} 
\usepackage{graphicx}
\usepackage{textcomp}
\usepackage{color, colortbl}
\usepackage[dvipsnames]{xcolor}
\def\BibTeX{{\rm B\kern-.05em{\sc i\kern-.025em b}\kern-.08em
    T\kern-.1667em\lower.7ex\hbox{E}\kern-.125emX}}

\usepackage{epsfig}
\usepackage{amsmath}
\usepackage{amssymb}
\usepackage{multirow}
\usepackage{xfrac}
\usepackage{adjustbox}
\usepackage{color, colortbl}
\usepackage[dvipsnames]{xcolor}
\usepackage{comment}

\definecolor{lightcyan}{rgb}{0.88,1,1}
\definecolor{lightgreen}{rgb}{0.80,1.0,0.80}

\usepackage{algorithm}
\usepackage[]{algpseudocode}
\usepackage{algorithmicx}

\usepackage{tikz}
\newcommand*\circled[1]{\tikz[baseline=(char.base)]{
            \node[shape=circle,fill=orange,inner sep=2pt] (char) {\textcolor{white}{#1}};}}

\usepackage{esvect}
\usepackage{hyperref}    

\hypersetup{nolinks=true} 

\usepackage{amsmath,amssymb,amsfonts}
\usepackage{textcomp}
\def\BibTeX{{\rm B\kern-.05em{\sc i\kern-.025em b}\kern-.08em
    T\kern-.1667em\lower.7ex\hbox{E}\kern-.125emX}}
    
\begin{document}

\title{Adaptive-Gravity: \\ A Defense Against
Adversarial Samples\\
}




\author{\IEEEauthorblockN{Ali Mirzaeian\IEEEauthorrefmark{1}, Zhi Tian\IEEEauthorrefmark{1}, Sai Manoj P D\IEEEauthorrefmark{1}, Banafsheh S. Latibari\IEEEauthorrefmark{3}, Ioannis Savidis\IEEEauthorrefmark{2},
\\Houman Homayoun\IEEEauthorrefmark{3},
Avesta Sasan\IEEEauthorrefmark{3}}
\IEEEauthorblockA{\IEEEauthorrefmark{1}Department of Electrical and Computer Engineering, George Mason University
}
\IEEEauthorblockA{\IEEEauthorrefmark{2}Department of Electrical and Computer Engineering, Drexel University}
\IEEEauthorblockA{\IEEEauthorrefmark{3}Department of Electrical and Computer Engineering University of California, Davis}
{\{amirzaei, ztian1, spudukot\}@gmu.edu, is338@drexel.edu, \{bsaberlatibari, hhomayoun, asasan\}@ucdavis.edu}}

\maketitle
\begin{abstract}\label{sec:abs}
This paper presents a novel model training solution, denoted as Adaptive-Gravity, for enhancing the robustness of deep neural network classifiers against adversarial examples. We conceptualize the model parameters/features associated with each class as a \textit{mass} characterized by its centroid location and the spread (standard deviation of the distance) of features around the centroid. We use the centroid associated with each cluster to derive an anti-gravity force that pushes the centroids of different classes away from one another during network training. Then we customized an objective function that aims to concentrate each class's features toward their corresponding new centroid, which has been obtained by anti-gravity force. This methodology results in a larger separation between different \textit{masses} and reduces the spread of features around each centroid. As a result, the samples are pushed away from the space that adversarial examples could be mapped to, effectively increasing the degree of perturbation needed for making an adversarial example. We have implemented this training solution as an iterative method consisting of four steps at each iteration: 1) centroid extraction, 2) anti-gravity force calculation, 3) centroid relocation, and 4) gravity training. Gravity's efficiency is evaluated by measuring the corresponding fooling rates against various attack models, including FGSM, MIM, BIM, and PGD using LeNet and ResNet110 networks, benchmarked against MNIST and CIFAR10 classification problems. Test results show that Gravity not only functions as a powerful instrument to robustify a model against state-of-the-art adversarial attacks but also effectively improves the model training accuracy.
\end{abstract}

\begin{IEEEkeywords}
Adversarial Example, Convolutional Neural Network, Latent Space Separation\
\end{IEEEkeywords}

\section{Introduction and Background}\label{sec:introduction}
Despite the evolution in neural networks model structures and their performance, it is illustrated that CNNs and DNNs are prone to adversarial attacks through simple perturbation of their input images \cite{fgsm, bim, mim, deepfool, dkdAli, CodeBridged}. These algorithms have demonstrated how easily the normal images can be perturbed by adding a small amount of noise to degrade the performance of neural networks. The vulnerability of deep neural networks to adversarial examples was first investigated in \cite{intriguing}. Since this early work, many new algorithms for generating adversarial examples and various solutions for defending against these attacks are proposed. 

Adversarial sample is introduced as an optimization problem, mathematically defined as follows \cite{intriguing}:
\begin{equation}\label{lbfgs}
    \begin{aligned}
        \operatorname*{argmin}_\epsilon f(x+\epsilon)=t \,\,\,\,\,\,\,\,
        s.t.\begin{cases}
        (x+\epsilon) \in D,\\
        f(x+\epsilon) \neq f(x)
        \end{cases}
    \end{aligned}
\end{equation}

In this optimization problem, $f$ is a classifier that maps image pixel vectors $x$ to a discrete k-label set $t$, i.e., $f: \mathbb{R}^m \rightarrow \{1 ... k\}$. The goal of this optimization formula is to find the minimum perturbation $\epsilon$, such that by applying it to the original data sample $x$, the under-attack machine learning model 
$f$ misclassifies the perturbed sample $x+\epsilon$ as the target class $t$, $f(x+\epsilon) = t$. 
The obtained perturbed sample $x+\epsilon$ also needs to remain in the acceptable input domain i.e., $D \in [0,1]^m$. In Szegedy and et al. \cite{intriguing}, this problem was solved using LBFGS algorithm. Although their offered solution is effective, it is a time-extensive process to achieve the adversarial perturbation. 

In \cite{fgsm}, Goodfellow and et al, introduced Fast Gradient Sign Method (FGSM) that, unlike LBFGS, was fast and effective. However, it perturbs all the input pixels for obtaining the adversarial example. But only a subset of input pixels can be found that has a similar effect and at the same time leads to a more imperceptible adversarial perturbation. Soon after FGSM, many algorithms like Basic Iterative Method (BIM)\cite{bim}, Momentum Iterative Method (MIM)\cite{mim}, and Projected Gradient Descent (PGD)\cite{minmax} have introduced. In which, through an iterative procedure a minimum amount of adversarial perturbation is generated that leads to a successful adversarial attack. 

In essence, adversarial examples could be constructed due to the underlying model's lack of adequate generality. By this intuition, many defenses against adversarial examples have been introduced that we illustrate some of the most notable ones as follow:

\textbf{Adversarial training} \cite{fgsm}: is an iterative procedure that at each iteration, the target model is being trained based on the training dataset. Different attacks are then applied to the model, and the extracted adversarial examples are added to the training dataset. This procedure continues till reaching an acceptable level of robustness. This method has two drawbacks: 1) it can only make the model robust against the assistant attacks; 2) It also increases the training time significantly. 

\textbf{Defensive Distillation}: In \cite{distillation}, distilling was used to propose the defense method. For defensive distillation, the second network is the same size as the first network \cite{distillation}. The main idea is to hide the gradients between the pre-softmax and softmax layers to make the attacker's job more difficult. However, it was illustrated in \cite{cw} that this defense could be broken by using the pre-softmax layer outputs in the attack algorithm and/or choosing a different loss function. 

\textbf{Gradient Regularization}: Input gradient regularization was first introduced in \cite{gradreg2} to improve the generalization of neural networks training by a double backpropagation method. The work in \cite{distillation} mentions the double backpropagation as a defense, and  \cite{gradreg} evaluates the effectiveness of this idea to train a more robust neural network. This approach intends to ensure that if there is a small change in the input, the change in Kullback-Leibler (KL) divergence between the predictions and the labels will also be small. However, this approach is sub-optimal because of the blindness of the gradient regulation. 
    
\textbf{Adversarial Detection}: Another approach is to detect adversarial examples before feeding them to the network \cite{detection, detection2}. \cite{detection} tries to find a decision boundary to separate adversarial and clean inputs. \cite{detection2} deploys the fact that the perturbation of pixel values by adversarial attack alters the dependence between pixels. By modeling the differences between adjacent pixels in natural images, deviations due to adversarial attacks can be detected.

\textbf{Autoencoders}: Authors in \cite{comparative} analyze the use of normal and denoising autoencoders as a defense method against adversarial samples. \cite{magnet}, uses a two-level module and uses autoencoders to detect and reform adversarial images before feeding them to the target classifier. However, this method may change the clean images and add a computational overhead to the whole defense-classifier module. To improve the method introduced in \cite{magnet},  the work in \cite{sabokrou2019self}  presents an efficient auto-encoder with a new loss function, which was learned to preserve the local neighborhood structure on the data manifold.

\begin{figure}[h]
  \centering
  \includegraphics[width=\columnwidth]{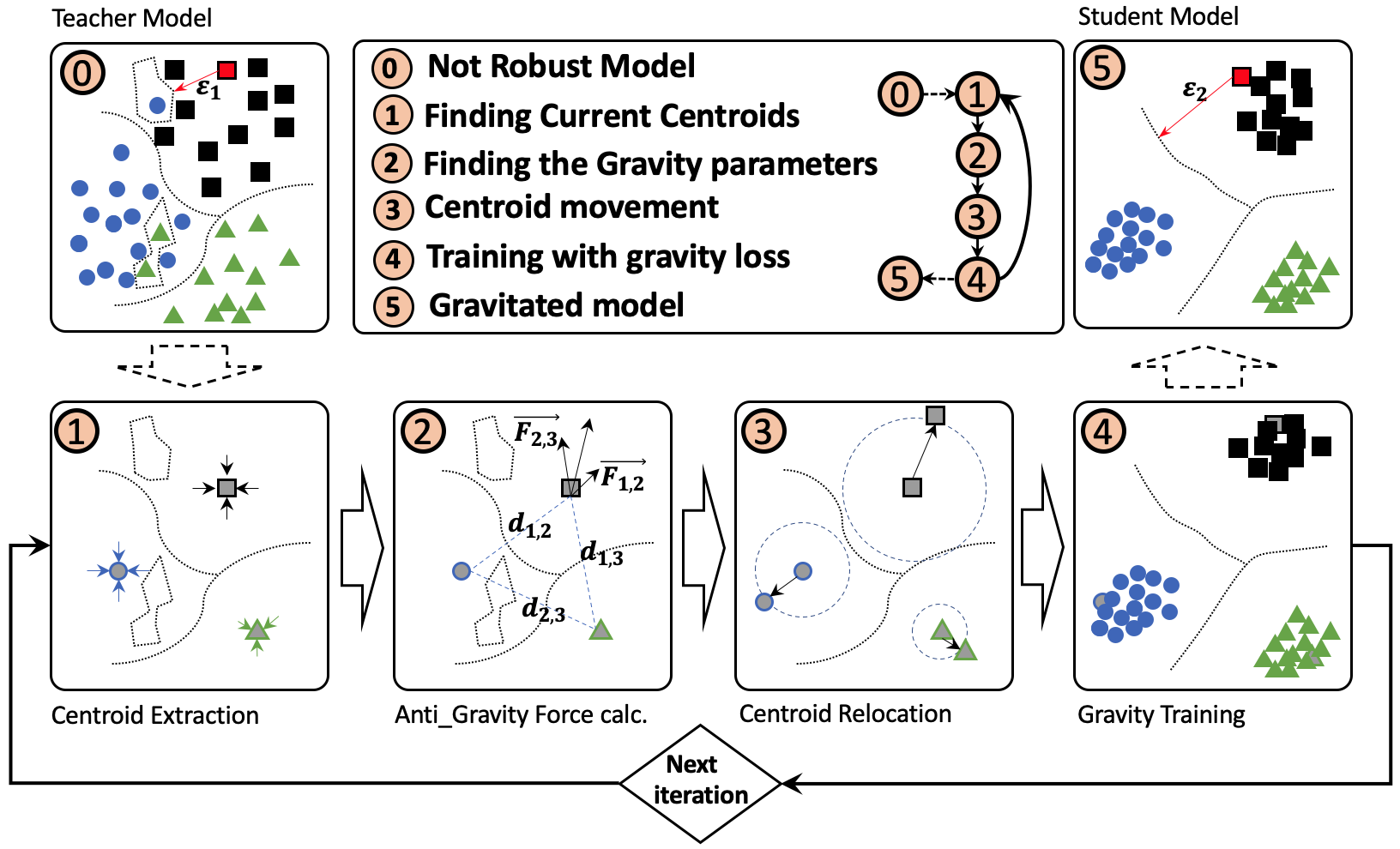}
  \caption {Gravitation process for robustifying a given model, indicated with step zero, through an iterative procedure, showed with steps one to four, to generate a  resilience model against adversarial attacks, showed with step five. }
 \label{gravity}
\end{figure}

\section{Adaptive-Gravity: Proposed Adversarial Defense }\label{sec:gravity}
In the previous sections, we have investigated different attacks and defenses related to the adversarial examples. One of the primary reasons for the attacks' existence is the low generality of the underlying model, i.e., the underlying model has trained on a dataset that is a subset of all the feasible data samples in the input domain. In this section, we present the proposed technique, termed Adaptive-Gravity, as an effective method that increases the underlying model's generality and simultaneously preserves (or increases) the accuracy of the ML classification model. Further, we combine the proposed gravity method with adversarial training to enhance the model robustness against adversaries. 
An abstract view of the proposed procedure is depicted in Fig. \ref{gravity}. Let's assume that Fig. \ref{gravity}-\circled{0} shows a latent space of a hypothetical model with three categories, indicated by black squares, blue circles, and green triangles. Due to the inherent high non-linearity of the model, there is a boundary of the class blue-circles inside the boundary of black-squares, making it much easier for an adversary to find a small adversarial perturbation $\epsilon_1$ which can lead to an imperceptible adversarial example.
  
In order to make it much harder for the adversary to find a perturbation, Adaptive-Gravity employs an iterative procedure comprised of four steps which are marked with \circled{1} to \circled{4} in Fig. \ref{gravity}. By following these steps a robust model can be obtained as shown in Fig. \ref{gravity}-\circled{5}. These steps will be elaborated on in this section. With this resultant model, the adversary needs a much larger perturbation $\epsilon_2$ to deceive the underlying model to misclassify the sample marked with the red rectangle in Fig. \ref{gravity}-\circled{5} compared with Fig. \ref{gravity}-\circled{0} i.e., $\epsilon_2 \texttt{>>} \epsilon_1$, which means the resilience of the gravitated model is much higher compared to the non-robust model.

\begin{figure}[h]
  \centering
  \includegraphics[width=0.8\columnwidth]{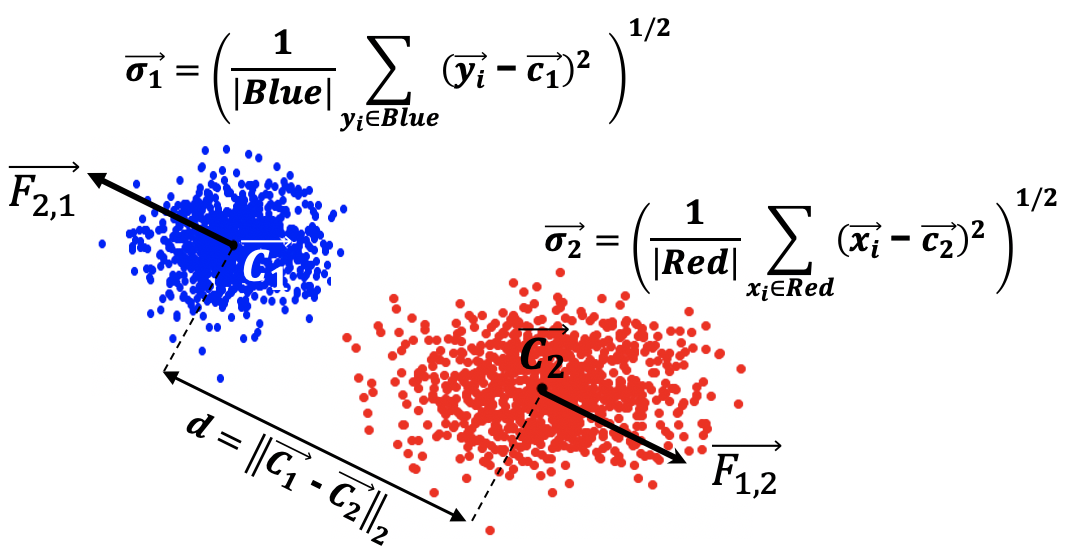}
  \caption {Definition of parameters in a model with two classes in which G is gravitization constant, $\vec{\sigma_1}$ and $\vec{\sigma_2}$ are the standard deviation of class 1 and class 2 respectively, $d$ is the $L_2$ norm distance between two centroids $c_1$ and $c_2$ which are associated to class 1 and class 2 respectively }
 \label{param-def}
\end{figure}

To understand the intuition behind Adaptive-Gravity, we first illustrate the concept 
of traditional gravity force between two masses $m_1$ and $m_2$ which are at distance of $d$ from each other as $\overrightarrow{F}_{m_1,m_2} =  G\frac{m_{1}m_{2}}{d^{2}}$, in which $G$ is constant \cite{newton1833philosophiae}. In this equation, the gravity force $\overrightarrow{F}_{m_1,m_2}$ increases linearly by increasing the mass of either $m_1$ or $m_2$, or it increases quadratically by reducing the distance between these two masses. With a similar intuition to the gravity force between two masses, we analogize each class as a mass at the centroid of that particular class. The standard deviation of each class, denoted by $\sigma$, is considered as the size of the mass. The $L_2$ norm between centroids of classes depicts the distance between their corresponding latent spaces. These parameters are shown in Fig. \ref{param-def} for two classes with a different distribution of their latent spaces. By defining these parameters, we formulate the anti-gravity force between two classes as described in Eq. \eqref{equ:repulsion} below. 
\vspace{-4pt}
\begin{equation}
    \overrightarrow{F}_{1,2} = \frac{\overrightarrow{\sigma}_{1} \overrightarrow{\sigma}_{2}}{d^{2}}(\overrightarrow{c_{1}}-\overrightarrow{c_{2}})
    \label{equ:repulsion}
\end{equation}

The anti-gravity force shows the direction and magnitude of the force that classes impose on each other to keep the different classes away from themselves. This formula serves the following purposes: 1) those classes that have a lower $L_2$ distance have a larger anti-gravity force, i.e., those classes that are already far apart from each other, having a higher $L_2$ distance,  have a minimum effect on other classes.  Conceptually, those classes that are closer to each other are more vulnerable to adversarial attacks. 2) Classes with a bigger mass (larger standard deviation) are more susceptible to adversarial attacks. Because samples of these classes are more likely to locate inside other classes' boundaries. This, in turn, lowers the trained model's generality and, consequently, makes it more vulnerable against adversarial attacks.

By this definition, Adaptive-Gravity iterates over four steps 1) Centroid Extraction, Fig. \ref{gravity}-\circled{1}, 2) Anti-Gravity force calculation, Fig. \ref{gravity}-\circled{2}, 3) Centroid relocation, Fig. \ref{gravity}-\circled{3}, 4) Gravity Training, Fig. \ref{gravity}-\circled{4}.


\subsection{Centroid Extraction}\label{extraction}
At this step, Fig. \ref{gravity}-\circled{1}, the mean of the embedded features in each class is considered as the corresponding centroid for that particular class. In this paper, we assume that each class has only one centroid, which can be obtained by averaging each class's embedded features. It is possible to consider more than one centroid for each class by employing the K-Means algorithm to locate multiple centroids for each class.  The obtained centroid for $i^{th}$ class  is 
denoted by $\overrightarrow{c_i} = \frac{1}{M_i}\sum\limits_{j=1}^{ M_i} \overrightarrow{x_{j}}$ in which $M_i$ is the cardinality of the $i^{th}$ class and $\overrightarrow{x_j}$ is the $j^{th}$ member of the $i^{th}$ class.  Each centroid is considered as a representative for a class that the whole mass of that class is concentrated on that centroid. Because all the steps of Gravity rely on the centroids of classes, the proper centroids can significantly change the performance of Gravity. 

\subsection{Anti-Gravity Force}\label{parameter}
The second step, Fig. \ref{gravity}-\circled{2}, is to obtain the total force on the i$^{th}$ centroid ($\overrightarrow{c_i}$) on a model with $N$ classes. According to Eq. \eqref{equ:repulsion}, the anti-gravity force $\overrightarrow{F}_{i,j}$ between two classes $i$ and $j$ is determined by three sets of quantities: $\overrightarrow{\sigma_{i}}$, $d_{ij}$, and $\overrightarrow{c_{i}}$. The standard deviation of the latent space related to i$^{th}$ class, $\phi_i$, is obtained by $\overrightarrow{\sigma_{i}}=\sqrt{\frac{1}{M_i}\sum\limits_{j=1}^{ M_i} \bigg(\overrightarrow{x_{j}}-\overrightarrow{c_{i}}\bigg)^2}$ in which $\overrightarrow{x_{i}} \in \phi_i$ and $M_i$ is the cardinality of the $i^{th}$ class.  The $L_2$ distance between $\overrightarrow{c_i}$ and the rest of the centroids ($\overrightarrow{c_j}$) is calculated by $d_{ij}=\parallel\overrightarrow{c_i}-\overrightarrow{c_j}\parallel_2$. 
Lastly, the total anti-gravity force on $\overrightarrow{c_i}$ is calculated by summing up
all the anti-gravity forces on this centroid as shown in Eq. \eqref{equ:repulsion-force}.

\begin{equation}
    \overrightarrow{F_{i}}=\sum_{j=1,j\neq i}^{N}\overrightarrow{F_{i,j}}=\sum_{j=1,j\neq i}^{N}\frac{\overrightarrow{\sigma_{i}}\overrightarrow{\sigma_{j}}}{d_{ij}^{2}}(\overrightarrow{c_{i}}-\overrightarrow{c_{j}})
    \label{equ:repulsion-force}
\end{equation}

The anti-gravity force for each centroid is different from others, meaning centroids are forced to move in different directions with different step sizes. Those centroids that are far from other centroids are subject to a weaker anti-gravity force than those centroids that are close to each other. That is because those classes that are close are more susceptible to adversarial attacks, so those should be separated more aggressively. 

\begin{figure}[h]
  \centering
  \includegraphics[width=\columnwidth]{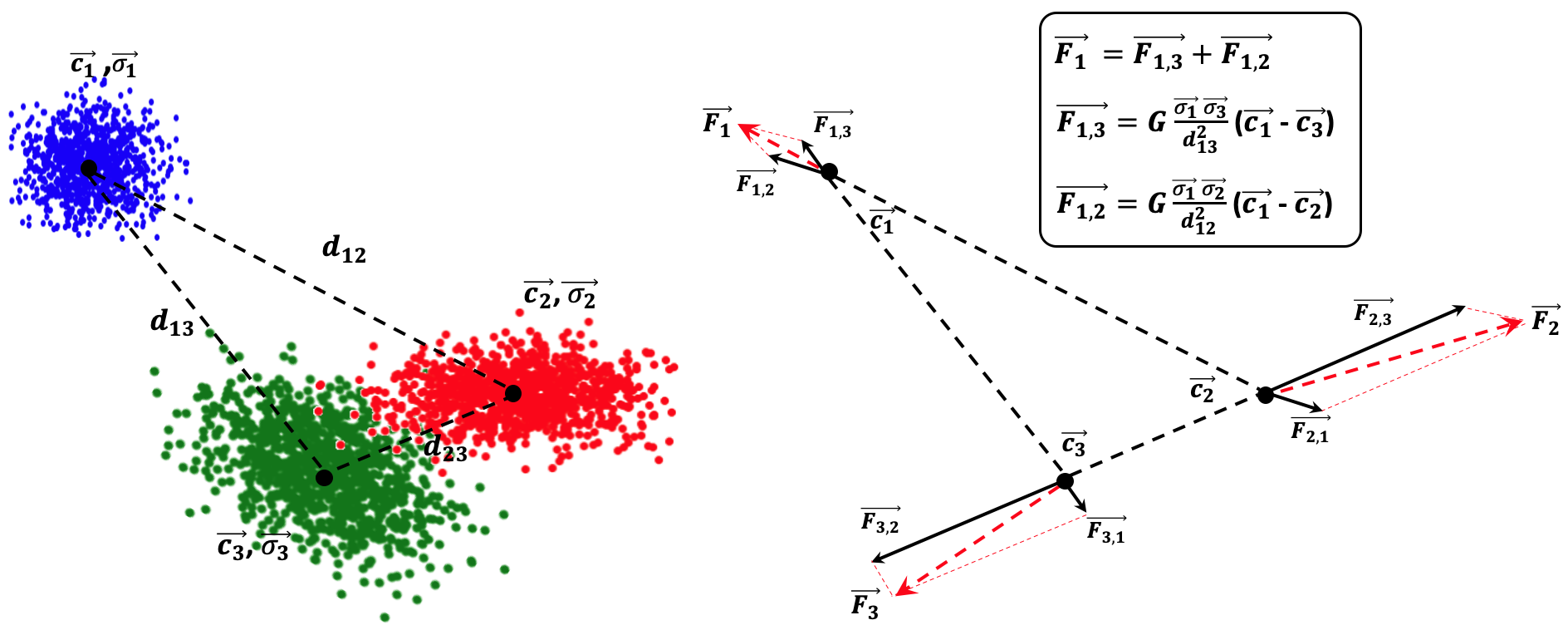}
  \caption {Total anti-gravity forces $\vec{F_{1}}$, $\vec{F_{2}}$,
  $\vec{F_{3}}$ corresponding to each one the classes with centroids 
  $\vec{c_{1}}$, $\vec{c_{2}}$, $\vec{c_{3}}$ and 
  standard deviations $\vec{\sigma_1}$, $\vec{\sigma_2}$, $\vec{\sigma_3}$, respectively.
  }
 \label{repulsion}
\end{figure}

For instance, consider a simple example of three classes with centroids $\overrightarrow{c_{1}}, \overrightarrow{c_{2}}, \overrightarrow{c_{3}}$ and standard deviations $\overrightarrow{\sigma_1}, \overrightarrow{\sigma_2}, \overrightarrow{\sigma_3}$ which are located at $L_2$ distances $d_{12}, d_{13}, d_{23}$ from each other as shown in Fig. \ref{repulsion}-left. In Fig. \ref{repulsion}-right, the anti-gravity force at each centroid has been shown with a red arrow. As shown,  the anti-gravity between classes two and three is much larger than the anti-gravity force toward class one. This is because classes two and three are close to each other and have a larger mass, while class one is far apart from the two others and has a smaller mass. Consequently, class one imposes the minimum anti-gravity force toward the other two classes, i.e., the other two classes have the minimum anti-gravity force toward class one.

\subsection{Centroid Relocation}
At this step, Fig. \ref{gravity}-\circled{3}, the obtained centroids in Section \ref{extraction} are relocated in the direction of the total anti-gravity forces obtained in section \ref{parameter}. Because there is no constraint on the magnitude of each centroid's total anti-gravity forces, the new centroids' locations can be too far from their current locations. So it is hard for the underlying model to get trained to converge the latent spaces of each class to their corresponding new centroid, which leads to a dramatic drop in the model's accuracy. To prevent this phenomenon, we define a normalized anti-gravity force $\overrightarrow{\alpha_i}$  for each class that constraints the maximum distance of the next centroid of a class to its current location.  $\overrightarrow{\alpha_i}$ is a vector in the same direction of $\overrightarrow{F_i}$ with a hyper-parameter $G$ to adjust the magnitude of it.  
This procedure is illustrated in Alg. \ref{centroid-modification}.


\begin{algorithm}[b]
\caption{Centroid Modification}
\label{centroid-modification}
\begin{algorithmic}[1]
\scriptsize

    \State \textbf{Inputs}:\textcolor{black}{$\overrightarrow{F}$: list of anti-gravity forces, $G$: constant, $C$: list of classes' centroid}
	\State \textcolor{black}{$F_M = max(|\overrightarrow{F}|_2)$}
	\For{($i=1$; $i \le N$ ; $i\texttt{+}\texttt{+}$)}
	    \State $\overrightarrow{\alpha_i} = G*(\overrightarrow{F}_i/F_M)$
	    \State $\overrightarrow{c}_{i}^{k+1} = \overrightarrow{\alpha_i} \texttt{+} \overrightarrow{c}_{i}^{k}$
    \EndFor
\end{algorithmic}
\end{algorithm}

In Alg. \ref{centroid-modification}, the inputs are anti-gravity forces $\overrightarrow{F}$, Gravity constant $G$, and the list of classes' centroids $c$. In line 2, the anti-gravity force that has the largest magnitude is selected then as in line 4, the normalized anti-gravity force $\overrightarrow{\alpha_i}$ for each one of the centroids is calculated, and finally, in line 5, each centroid is updated by the calculated $\overrightarrow{\alpha_i}$. With this algorithm, each centroid at the $k^{th}$ iteration of Gravity i.e., $\overrightarrow{c}_{i}^{k}$, can be relocated within the $L_2$ distance range $(0, G]$ in the same direction of $\overrightarrow{\alpha_i}$.

Using this simple method, Adaptive-Gravity adds a priority measure to each centroid to indicate how aggressively each centroid should move. Adaptive-Gravity assigns a higher priority to separate classes close to one by using a maximum step size $G$, while spending less effort in separating classes that are far apart. For example, in Fig. \ref{repulsion}, $\overrightarrow{F_1}$ has the lowest magnitude, so at the next iteration $\overrightarrow{c_1}$ has the smallest relocation compared to other two centroids $\overrightarrow{c_2}$  and $\overrightarrow{c_3}$ . 
\subsection{Gravity Training}
Gravity training, Fig.\ref{gravity}-\circled{4}, is an iterative procedure that performs in a teacher/student fashion. The student model has the same structure as the teacher model. During this process, a student model is trained such that its accuracy follows the teacher's accuracy but covers the teacher model's weakness against adversarial attacks. 
\subsubsection{Gravity Force}
So far, the formulated training process pushes the centroids of different clusters away from one another.  To further increase the separation, we also need to reduce each mass's spread (standard deviation of the distance between features and centroid in each class). By reducing the mass of the classes, they become more concentrated toward their centroids. It needs to be noted that, by decreasing the standard deviation of classes, the latent spaces of classes are less likely to overlap, and this leads to a model with a higher generality that is less vulnerable toward adversarial examples.

For this purpose, gravity force has been defined to reduce the standard deviation of classes. Expressly, we represent the gravity force as Mean Square Error (MSE) between each class's latent spaces and its corresponding centroids. Conceptually, minimizing the gravity force of a class leads to a more concentrated latent space around its centroid. To model the force of gravity in our optimization problem, we formulate the gravity force using the \textit{gravity loss function}, details of which are discussed in the next section.

\subsubsection{Gravity Loss Function}
At this step, the model is trained to minimize the gravity loss using the Eq. \eqref{eq:grloss} through an iterative process. This equation consists of two parts: 1) The first part that targets accuracy is annotated with L$_{ce}^{(k)}$ and defines the cross-entropy loss, Eq. \ref{eq:entropy}, between the ground truths $Y$ and the outputs of the student model i.e., $S(X_{m \times n}, Y)$, at the $k^{th}$ iteration. 2) The second part that targets security aims to minimize the gravity force, i.e., converging the latent space, $\phi_i$, toward their corresponding centroids $\overrightarrow{c_i}$. This structure is similar to the defensive distillation method discussed earlier. Despite the similar structure, Gravity uses two layers, the head, and tail of the teacher model, to transfer their knowledge to the student model's corresponding layers. 

\begin{equation}
    \small
    L^{(k)}_{Gr} = \underbrace{(1-\gamma_{{k-1}})L_{ce}^{(k)}}_{targets\, the\, accuracy} + \underbrace{\gamma_{{k-1}}(L^{(l=-1,k)} + L^{(l=i,k)})}_{targets \, the \, security}
    \label{eq:grloss}
\end{equation}
\normalfont


\begin{equation}
    \small
    L_{ce}^{(k)}=-\sum_{i=1}^{i=m \times n}{Y_i \log(S^k(X_i))}
    \label{eq:entropy}
\end{equation}
\normalfont

L$^{(l=-1,k)}$ and L$^{(l=i,k)}$ are associated to the student's head and tail objective function at the $k^{th}$ iteration and are defined as in Eq. \eqref{eq:ltail} with $l=-1$ and $l=i$,  respectively. In these equations, 
Mean Square Error (MSE) calculates the likelihood between the latent spaces of the head (or tail) layer of the student model, $S_{\phi_{m \times n}}^{(l=-1, k)}$, and the corresponding centroids of the teacher model, $S_{c_{m}}^{(l=-1, k-1)}$.
\begin{equation}
    \small
    L^{(l,k)}=\frac{1}{m \times n}\sum_{j=1}^{m}\sum_{f=1}^{n}\bigg(S_{c_j}^{(l, k-1)} - S_{\phi_{(f,j)}}^{(l, k)}\bigg)^2
    \label{eq:ltail}
\end{equation}
\normalfont

Several remarks are in order. First, at the first iteration, Adaptive-Gravity transfers as much as possible knowledge from the teacher model to the student model. But at the later iterations, Adaptive-Gravity focuses on securing the student model, i.e., removing those vulnerabilities transferred from the teacher model during the first iteration. Second, $\gamma_{i}$ contains the accuracy of the teacher model at each iteration. So if this accuracy is high, the student model relies more on the teacher model, which brought in the Gravity Loss by the second term. However, if the teacher model's accuracy is low, the student model relies on itself, which is the first term. Using this strategy, the student model adds more intelligence to its learning process to not follow the teacher model if performance degradation occurs.

\begin{table}
    \centering
    \caption{LeNet and Resnet-110 structures are used for training MNIST and CIFAR10 datasets, respectively. The Head and Tail layers for each model have been highlighted. System Configuration and training hyperparameters also have shown on the bottom.}
    \resizebox{.8\width}{!}{
        \begin{tabular}{|c|c|}
            \hline 
            \rowcolor{lightgray}
            \textbf{LeNet} & \textbf{ResNet-110}\tabularnewline
            \hline 
            \hline 
             
                \begin{tabular}{c}
                Conv(6, $5\times5$)\tabularnewline
                ReLu($2\times2$)\tabularnewline
                \end{tabular} & %
                \begin{tabular}{c}
                Conv(16, $3\times3$)+BN\tabularnewline
                ReLu($2\times2$)\tabularnewline
                \end{tabular}\tabularnewline
             
                \begin{tabular}{c}
                Conv(16, $5\times5$)\tabularnewline
                PReLu($2\times2$)\tabularnewline
                \end{tabular} & \bigg[%
                \begin{tabular}{c}
                Conv(16, $1\times1$)+BN\tabularnewline
                Conv(16, $3\times3$)+BN\tabularnewline
                Conv(64, $1\times1$)+BN\tabularnewline
                \end{tabular}\bigg]$\times12$\tabularnewline
            \cellcolor{lightcyan}
             \begin{tabular}{c}
             FC(120)
             Tail
             \end{tabular}& \bigg[%
                \begin{tabular}{c}
                Conv(32, $1\times1$)+BN\tabularnewline
                Conv(32, $3\times3$)+BN\tabularnewline
                Conv(128, $1\times1$)+BN\tabularnewline
                \end{tabular}\bigg]$\times12$\tabularnewline
             FC(84) & \bigg[%
                \begin{tabular}{c}
                Conv(64, $1\times1$)+BN\tabularnewline
                Conv(64, $3\times3$)+BN\tabularnewline
                Conv(256, $1\times1$)+BN\tabularnewline
            \end{tabular}\bigg]$\times12$\tabularnewline
            \cellcolor{lightgreen}
            FC(10) Head     &  \cellcolor{lightcyan} FC(1024) Tail \tabularnewline
              & \cellcolor{lightgreen} FC(10) Head\tabularnewline
    \hline 
    \hline
    \rowcolor{lightgray}
    \multicolumn{2}{|c|}{ \textbf{System Configuration and training hyper parameters} }\tabularnewline

    		\hline 
    		\multicolumn{2}{|p{1\linewidth}|}{OS: Red Hat 7.7, Pytorch: 1.3, AdverTorch: 0.2,  GPU: Nvidia Tesla V100, EPOCH: 100,
    		MNIST Batch Size: 64, CIFAR10 Batch Size:128, Optimizer: ADAM, MNIST learning rate: 1e-4, CIFAR10 learning rate: 5e-3 }\tabularnewline
    		\hline
    \end{tabular}}
    \label{structures}
\end{table}

\section{Experiments}\label{sec:experiments}
In order to assess the performance of the Adaptive-Gravity method we used two architectures LeNet \cite{lecun1998gradient} and ResNet-110 \cite{resnet}, see Table \ref{structures}, for training and evaluating on two datasets MNIST \cite{lecun1998gradient} and CIFAR10 \cite{krizhevsky2009learning}, respectively. In this section, we evaluate our results in three sections 1) performance of proposed gravity technique, the impact of Gravity on 2) White-box attacks, 3) Black-box attacks. The first section shows how effective Gravity separates the latent spaces of a target layer of an underlying model. Sections two and three show how resistant the gravitated model is against different adversarial examples in the white-box and black-box scenarios.

\begin{figure*}[h]
  \centering
  \includegraphics[width=1.6\columnwidth]{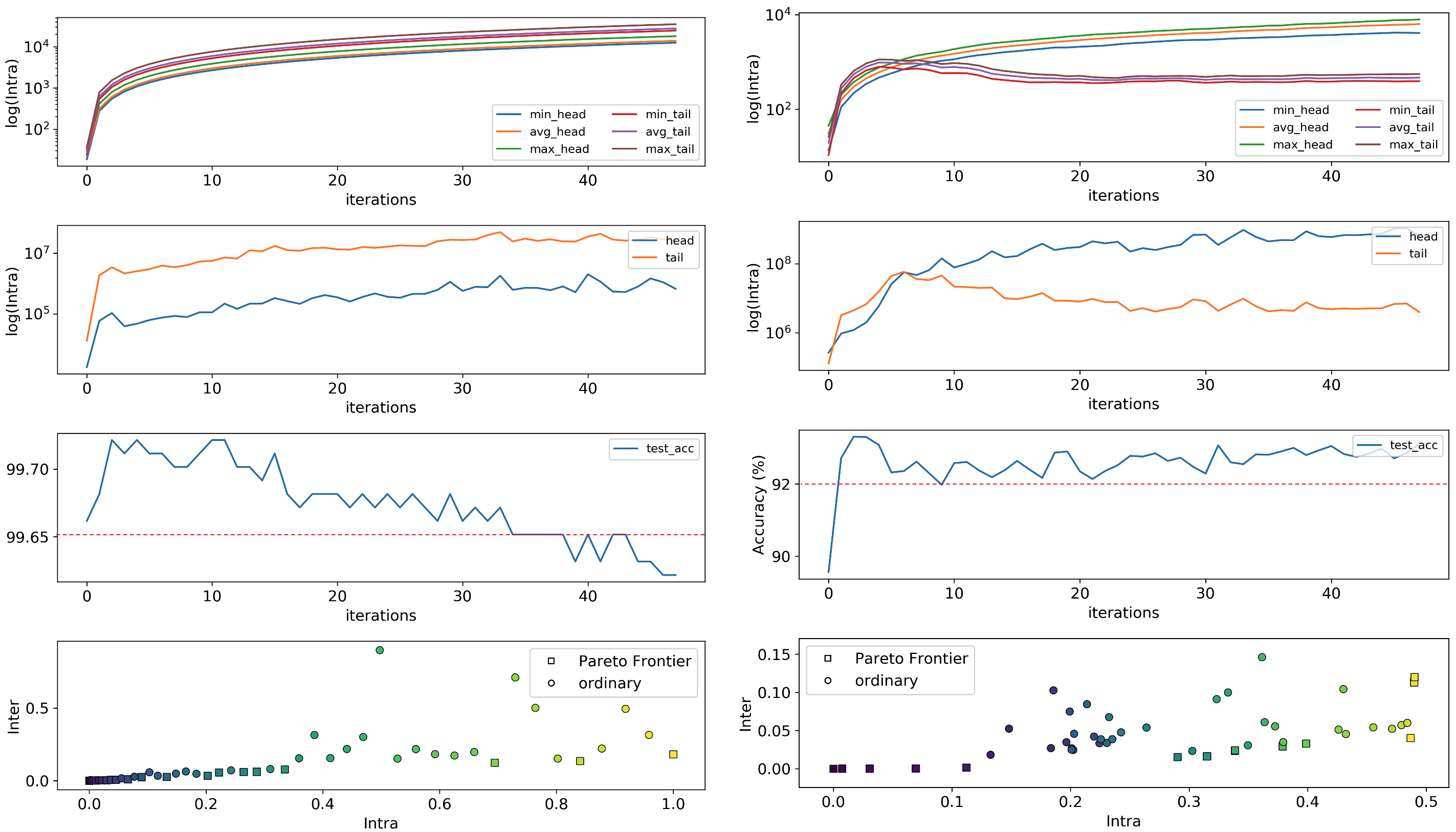}
  \caption {ICD (first row), ICC (second row), accuracy (thrid row), Pareto metrics of gravity during different iterations. The fourth row shows the relation between normalized ICC and ICD. In the fourth row, purple is related to the early iteration, and yellow is related to the 50$^{\mbox{th}}$ iteration. Left and right figures are related to the LeNet and  ResNet110 models trained on MNIST and CIFAR10 datasets, respectively.}
 \label{icc_icd}
\end{figure*}
\vspace{-5pt}

\subsection{Gravity Performance}
In this section, we measure three metrics 1) Inter-Class Convergence (ICC), that indicates how far, $L_2$ norm, are the latent spaces of the head or tail layer from their designated centroid at each iteration. ICC is obtained by summing masses of all classes, see Fig. \ref{param-def}, i.e., $\sum_{i=1}^{N}\epsilon_i$. 2) Intra-Class Divergence (ICD), measures the relative distance, $L_2$ norm, of the designated centroids for the head and tail layers at each iteration, see Fig. \ref{param-def}. ICD contains three values I) Average distance i.e., $1/N\sum_{i=1}^{N-1}\sum_{j=i+1}^{N} d_{ij}$, II) Maximum distance i.e., $\mbox{max}(d_{ij})$, and III) Minimum distance i.e., $\mbox{min}(d_{ij})$. In which $\mbox{d}_{ij}$ indicates the distance between centroids of $\mbox{i}^{th}$ and $\mbox{j}^{th}$ classes. 3) Evaluation accuracy,  that measures the student evaluation accuracy at each iteration.

Figure \ref{icc_icd} shows the evaluation of these three metrics during 50 iterations of Gravity for training the models ResNet110 and LeNet, see the table \ref{structures}, on two datasets CIFAR10 and MNIST, respectively. By studying Fig. \ref{icc_icd} we have the following observations: 

\begin{itemize}
\item The first row shows a different pattern for the ICD metric associated with the head and tail layers of LeNet compared with ResNet110. In the LeNet, the ICD of the head and tail layers show a similar behavior by increasing in a logarithmic form. Meaning, in the early iterations of Gravity, the ICD increases with a larger magnitude than later iterations. Unlike LeNet, in the ResNet110 after the 5$^{\mbox{th}}$ iteration, the ICD of the tail layer comes under the head layer. One interpretation is that in the LeNet, Gravity successfully moves the centroids further apart because MNIST is a simple dataset. Increasing the ICD of the head layer has no contradictory impact on the ICD of the tail layer. However, CIFAR10 is a much more complex dataset, and consequently, after 5$^{\mbox{th}}$ iteration increasing the ICD of the head layer leads to decreasing the ICD of the tail layer.

\item The second row, associated with the ICC metric for the tail and head layers, shows different behavior regarding the LeNet and ResNet110. For the LeNet, ICC increases rapidly at the early iterations, but later, this growth is diminishing in both head and tail layers. For the ResNet110, the ICC of the head layer continually increases (more rapidly at the early iterations), but the ICC of the tail layer rapidly increases till the 7$^{\mbox{th}}$ iteration then start a diminishing decline. 

\item The third row, which is related to the evaluation of the student model's accuracy, confirms that Gravity can increase the accuracy of the underlying models while separating the latent spaces of the student's head and tail layers. Noted that the reported accuracy at the 0$^{\mbox{th}}$ iteration indicates the original accuracy of the model we wanted to harden it. 
One reason for improving accuracy is that Gravity helps increase the generality of the models, i.e., models show a higher accuracy at the evaluation phase. 
\end{itemize}

\begin{figure}
  \centering
  \includegraphics[width=\columnwidth]{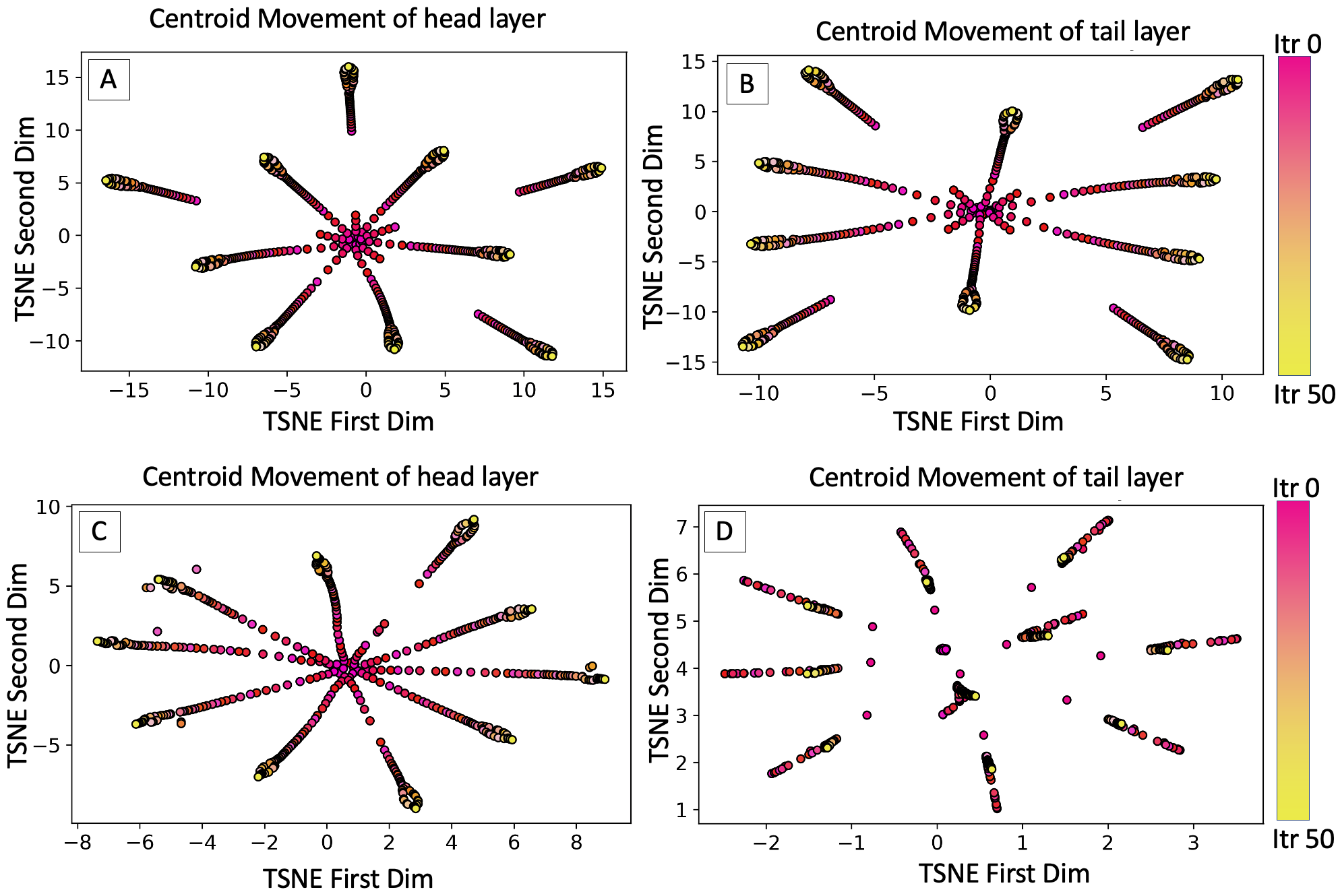}
  \caption {(A), (B) shows the LeNet centroid movement of the head and tail layers with $G_{head}$ = 100 and $G_{tail}$ = 200 on MNIST dataset. (C), (D) shows the ResNet110 centroid movement of the head and tail layers with $G_{head}$ = 200 and $G_{tail}$ = 300 on CIFAR10 dataset. Iterations are color-coded with red as the 0$^{\mbox{th}}$ iteration to yellow as the 50$^{\mbox{th}}$ iteration.}
 \label{mnist_cifar_lss}
\end{figure}

A secure model (against adversarial input perturbation attacks) should resist adversarial examples and generate the correct classification outcome maintaining a high classification accuracy for both adversarial and normal inputs.  By increasing ICD and decreasing ICC, and preserving (or increasing) the accuracy, we expect the underlying model's robustness to increase. However, from our observations presented in Fig. \ref{icc_icd}, these conditions are not always satisfied during different iterations. Hence we need to embrace a strategy to select one iteration of Gravity that is the most suitable for the robustifying task. For this purpose, we establish an acceptable threshold for accuracy at the first step, which is shown with dash lines in the third row of fig. \ref{icc_icd}. In the next step, we measure the relation between the normalized ICC which is defined as  $\scriptsize\frac{\mbox{MinICC}_{\mbox{head}} \times \mbox{MinICC}_{\mbox{tail}}}{\mbox{Max}(\mbox{MinICC}_{\mbox{head}}) \times \mbox{Max}(\mbox{MinICC}_{\mbox{tail}})}$, and the normalized ICD which is defined as $\scriptsize\frac{\mbox{ICD}_{\mbox{head}} \times \mbox{ICD}_{\mbox{tail}}}{\mbox{Max}(\mbox{ICD}_{\mbox{head}}) \times \mbox{Max}(\mbox{ICD}_{\mbox{tail}})}$, for those iterations resulting in an accuracy level above the threshold.  In the third step, we obtain the Pareto-front between those iterations that satisfying the threshold requirement. Noted that each Pareto-front is associated with one iteration, so in the fourth step, between these Pareto-fronts, we pick one iteration that leads to the best resistance against adversarial attack. Here we used the PGD attack as one of the most powerful first-order adversarial attacks. Finally, The model parameters generated in the selected iteration are then used in the student model.

We set the threshold to 0.9965 for the MNIST dataset and 0.92 for the CIDAR10 dataset, as shown with the red-dash line in Fig. \ref{icc_icd} third row. Then we obtained the Pareto-fronts for those selected iterations as shown in Fig. \ref{icc_icd} fourth row. Between all the Pareto-fronts, the 10$^{\mbox{th}}$ and 8$^{\mbox{th}}$ iterations show the most resilience against PGD attack for the LeNet and ResNet110, respectively. So for the next evaluations, we use the parameters associated with these iterations. 




Figures \ref{mnist_cifar_lss}-(A) and (B) illustrate the trajectory of the centroid movement of the latent spaces associated with the head and tail layers of LeNet, marked with blue and green in Table \ref{structures}, on the MNIST dataset. In these figures, the centroids of classes at each iteration are shown with a different color. For example, the red color is associated with the ten centroids at the first iteration, and similarly, the yellow ones are corresponding to 50$^{\mbox{th}}$ iteration. This figure confirms that by increasing the number of iterations at the Adaptive-Gravity, the classes and subsequently their related centroids are moving away from each other. Figure \ref{mnist_cifar_lss} captures similar information for the head and tail layers of ResNet110 that was trained on the CIFAR10 dataset. Unlike LeNet, the trajectory of centroid movement of head and tail layers of ResNet110 shows different behavior, as explained earlier in this section.

\begin{table*}
	\caption{Comparison of Gravity method to other defense methods on CIFAR10 dataset. For Gravity, we reported results without adversarial training (Gravity) and with adversarial training using FGSM (Gravity$_f$) and PGD (Gravity$_p$) attacks.} 
	\label{table:wb:cifar}
	\centering
	\resizebox{.75\width}{!}{
	\begin{tabular}{|c|c|c|c|c|c|c|c|c|c|c|c|}
		\hline 
		Attacks               & Params.                    & Baseline & AdvTrain\cite{kurakin2016adversarial} & Ye et al. \cite{yu2018interpreting} & Ross et al. \cite{ross2017improving} & Pang et al.\cite{pang2019improving} & Madry et al. \cite{madry2017towards} & Mustafa et al. \cite{mustafa2019adversarial} & Gravity & Gravity$_f$ & Gravity$_p$\tabularnewline 
		\hline 
		\hline 
		No Attacks            & -                          &   89.0       & 84.5            & 83.1              & 86.2                & 90.6               & 87.3                 & 90.5                   &     \textbf{93.0}    &      91.9       & 91.32 \tabularnewline            
		\hline 
		\multirow{2}{*}{FGSM} & $\epsilon$=0.02              &    74.0      & 44.3            & 48.5              & 39.5                & 61.7               & 71.6                 & 72.5                   &    90.5     &      \textbf{90.94}       & 89.57 \tabularnewline            
		                      & $\epsilon$=0.04              &    46.0      & 31.0            & 38.2              & 20.8                & 46.2               & 47.4                 & 56.3                   &    88.57     &       \textbf{90.12}      & 87.79\tabularnewline            
		\hline 
		\multirow{2}{*}{BIM}  & $\epsilon$=0.01              &      82.57    & 22.6            & 62.7              & 19.0                & 46.6               & 64.3                 & 62.9                   &    \textbf{92.45}     &      91.79       & 91.31 \tabularnewline            
		                      & $\epsilon$=0.02              &     77.11     & 7.8             & 39.3              & 6.9                 & 31.0               & 49.3                 & 40.1                   &     \textbf{92.28}    &      91.77       & 91.25\tabularnewline            
		\hline 
		\multirow{2}{*}{MIM}  & $\epsilon$=0.01              &     76.41     & 23.9            & -                 & 24.6                & 52.1               & 61.5                 & 64.3                   &    90.46     &      \textbf{91.38}       &  90.66 \tabularnewline            
		                      & $\epsilon$=0.02              &     53.93     & 9.3             & -                 & 9.5                 & 35.9               & 46.7                 & 42.3                   &   88.58      &     \textbf{90.96}        & 89.67\tabularnewline            
		\hline 
		\multirow{2}{*}{PGD}  & $\epsilon$=0.01              &    82.8      & 24.3            & -                 & 24.5                & 48.4               & 67.7                 & 60.1                   &     91.12    &      \textbf{91.7}       & 91.09 \tabularnewline            
		                      & $\epsilon$=0.02              &     74.38     & 7.8             & -                 & 8.5                 & 30.4               & 48.5                 & 39.3                   &    86.21     &       \textbf{91.29}      & 90.47 \tabularnewline            
		\hline 
		
		\multirow{3}{*}{C\&W}  & c=0.001              &    52.3      & 67.7           & 82.5                 & 72.2                & 80.6               & 84.5                 & 91.3                   &     90.82    &      \textbf{91.75}       & 91.09 \tabularnewline            
          & c=0.01              &     47.38     & 40.9            & 62.9                 & 47.8                 & 54.9               & 65.7                & 73.7                  &    85.24     &       87.27      & \textbf{89.48} \tabularnewline  
        & c=0.1              &     10.38     & 25.4            & 40.7                & 19.9                 & 25.6               & 47.9                & 60.5                  &    83.23     &       \textbf{85.29}      & 84.17 \tabularnewline 
		\hline 
	\end{tabular}}
\end{table*}

\begin{table*}
	\caption{ Comparison of Gravity method to other defense methods on MNIST dataset. For Gravity, we reported results without adversarial training (Gravity) and with adversarial training using FGSM (Gravity$_f$) and PGD (Gravity$_p$) attacks.} 
	\label{table:wb:mnist}
	\centering
	\resizebox{.8\width}{!}{
	\begin{tabular}{|c|c|c|c|c|c|c|c|c|c|c|}
		\hline 
		Attacks                    & Params.       & Baseline & AdvTrain\cite{kurakin2016adversarial} & Ye et al. \cite{yu2018interpreting} & Ross et al. \cite{ross2017improving} & Pang et al.\cite{pang2019improving} &  Mustafa et al. \cite{mustafa2019adversarial} & Gravity & Gravity$_f$ & Gravity$_p$\tabularnewline 
		\hline 
		\hline 
		No Attacks                 & -             &     98.63     & 99.1            & 98.4              & 99.2                & 99.5               & 99.5                  &    \textbf{99.69}     &      99.68       & 99.59\tabularnewline            
		\hline 
		\multirow{2}{*}{FGSM}      & $\epsilon$=0.1  &     91.23     & 73.0            & 91.6              & 91.6                & 96.3               & 97.1                  &     91.53      &       97.2      &  \textbf{97.67}\tabularnewline            
		                           & $\epsilon$=0.2  &     9.7     & 52.7            & 70.3              & 60.4                & 52.8               & 70.6                  &    70.68    &     68.09    & \textbf{73.06}\tabularnewline            
		\hline 
		\multirow{2}{*}{BIM}       & $\epsilon$=0.1  &     93.72     & 62.0            & 88.1              & 87.9                & 88.5               & 90.2                  &    98.35     &      98.67       & \textbf{98.72}\tabularnewline            
		                           & $\epsilon$=0.15 &     22.2     & 18.7            & 77.1              & 32.1                & 73.6               & 76.3                  &    97.54     &     98.56        & \textbf{98.66}\tabularnewline            
		\hline 
		\multirow{2}{*}{MIM}       & $\epsilon$=0.1  &     90.9     & 64.5            & -                 & 83.7                & 92.0               & 92.1                  &    97.07      &      97.28       & \textbf{98.61}\tabularnewline            
		                           & $\epsilon$=0.15 &     7.6     & 28.8            & -                 & 29.3                & 77.5               & 77.7                  &    81.49     &     \textbf{84.44}        & 83.34\tabularnewline            
		\hline 
		
		
		\multirow{2}{*}{PGD}       & $\epsilon$=0.1  &    20.15      & 62.7            & -                 & 77.0                & 82.8               & 83.6                  &    94.95     &      \textbf{98.46}       & 98.45\tabularnewline            
		                           & $\epsilon$=0.15 &     6.99     & 31.9            & -                 & 44.2                & 41.0               & 62.5                  &    68.26     &      96.97       & \textbf{97.26}\tabularnewline            
		\hline

		\multirow{3}{*}{C\&W}       & $c$=0.1  &    60.15      & 71.1            & 89.2                 & 88.1               & 97.3               & 97.7                 &    96.55     &      97.3       & \textbf{98.05}\tabularnewline            
		    & $c$=1 &     36.99     & 39.2            & 79.1                 & 75.3               & 78.1               & 91.2                  &    95.6     &      97.87       & \textbf{97.92}\tabularnewline  
          & $c$=10 &     6.01     & 17.0            & 37.6                & 20.0                & 23.8               & 46.0                 &    85.26     &      \textbf{86.97}       & 85.96\tabularnewline 
		\hline 
				
	\end{tabular}}

\end{table*}

\subsection{Performance against White-Box Attacks}

In white-box attack the attacker has access to the trained model, i.e., $S(.)$, and the training dataset. Thus, by having these two components, we assess how many of the input samples can be misclassified by the adversary, i.e., the accuracy of the underlying model in the existence of different adversarial attacks indicates the robustness of the model.


The structure of the used model is shown in Table \ref{structures} in which colored rows show the location of the tail and head in the Adaptive-Gravity method. We also combined the Adaptive-Gravity with adversarial training techniques for boosting the model robustness. For adversarial training, we used two different attacks FGSM and PGD, and their results are reported under the names AdvTrain$_{\mbox{\scriptsize FGSM}}$ and AdvTrain$_{\mbox{\scriptsize PGD}}$, respectively. We have also compared the proposed Adaptive-Gravity technique's performance with the existing aforementioned adversarial defenses in Tables \ref{table:wb:cifar} and \ref{table:wb:mnist} on datasets CIFAR10 and MNIST, respectively.

By investigating the white-box scenario's result, we have observed that: 1) The accuracy of the gravitated model is higher than the baseline. This shows that Adaptive-Gravity has spaced the classes far apart, allowing for better generalization by simplifying the class mapping boundaries, resulting in the correct classification of near-boundary samples misclassified by the baseline model.  2) Combining adversarial training with Gravity has a trivial impact on the accuracy of the underlying model. This is because Adaptive-Gravity has already generalized the model, and combining more samples (adversarial samples) into the training dataset will not significantly change the model's performance. 3) Compared to the other defenses, Adaptive-Gravity shows a much better resilience against adversarial attacks.

\subsection{Adaptive-Gravity against Black-Box Attacks}
In The black-box attacks, the attacker only has access to the trained data and can only feed input samples to the model under attack and observe its output. For launching this scenario, we have assumed the attacker has chosen the VGG-19 model to generate adversarial examples and feed them to the model which has trained with the Adaptive-Gravity strategy. We report the results of a black-box attack on a trained model based on Adaptive-Gravity and adversarial training in Table \ref{table:bb}.

\begin{table}
    \scriptsize
    \centering
	\caption{Robustness of our model in black-box settings. $\epsilon_{\mbox{\scriptsize FGSM}}$, $\epsilon_{\mbox{\scriptsize BIM}}$, $\epsilon_{\mbox{\scriptsize MIM}}$, $\epsilon_{\mbox{\scriptsize PGD}}$  shows the equivalent $\epsilon$ for each one of the attacks FGSM, BIM, MIM, and PGD, that lead to the same PSNR in white-box settings. We also investigate the impact of combining Gravity with adversarial training.}
	\resizebox{.94\width}{!}{
	\begin{tabular}{|c|ccccc|}
		\hline 
		Training                  & \multicolumn{1}{c|}{No Attack} & \multicolumn{1}{c|}{FGSM} & \multicolumn{1}{c|}{BIM}  & \multicolumn{1}{c|}{MIM} & PGD\tabularnewline 
		\hline 
		\hline 
		\rowcolor{lightgray}
		\multicolumn{6}{|c|}{MNIST ($\epsilon_{\mbox{\scriptsize FGSM}}=0.225$,  $\epsilon_{\mbox{\scriptsize BIM}}=0.1$, $\epsilon_{\mbox{\scriptsize MIM}}=0.2$, $\epsilon_{\mbox{\scriptsize PGD}}=0.265)$}\tabularnewline
		\hline 
		Baseline                  &              99.3         &       39.2       &      20.15                    &             16.28               & 20.1\tabularnewline    
		\cline{1-1} 
		Gravity                   &              99.69                  &              87.33             &           94.95               &                 91.05             & 94.0 \tabularnewline    
		\cline{1-1} 
		$Gravity+AdvTrain_{\mbox{\scriptsize FGSM}}$ &               \textbf{99.76}                 &            \textbf{96.41}               & \textbf{97.92}                          &                \textbf{96.93}            & 97.0 \tabularnewline    
		\cline{1-1} 
		$Gravity+AdvTrain_{\mbox{\scriptsize PGD}}$  &                \textbf{99.76}                &             96.04              &                 97.69         &              96.52                &  \textbf{97.65} \tabularnewline    
		\hline 
		\rowcolor{lightgray}
		\multicolumn{6}{|c|}{CIFAR10 ($\epsilon_{\mbox{\scriptsize FGSM}}=0.01$, $\epsilon_{\mbox{\scriptsize BIM}}=0.01$, $\epsilon_{\mbox{\scriptsize MIM}}=0.01$, $\epsilon_{\mbox{\scriptsize PGD}}=0.03)$}\tabularnewline
		\hline 
		Baseline                  &              92.4                  &      83.8                     &           85.6               &     84.1                         & 85\tabularnewline    
		\cline{1-1} 
		Gravity                   &              \textbf{92.5}                  &    87.7                       &           88.9               &      88.2                        & 88.9\tabularnewline    
		\cline{1-1} 
		$Gravity+AdvTrain_{\mbox{\scriptsize FGSM}}$ &             91.81                   &     \textbf{89.45}                      &           \textbf{89.9}               &     \textbf{89.5}                        & \textbf{90.0} \tabularnewline    
		\cline{1-1} 
		$Gravity+AdvTrain_{\mbox{\scriptsize PGD}}$  &         92.4                       &       87.7                    &           88.6               &        87.9                      & 88.6 \tabularnewline    
		\hline 
	\end{tabular}}
	\label{table:bb}
	
\end{table}

\vspace{-5pt}

\section{Conclusion}
This paper introduces Adaptive-Gravity, a powerful training solution for hardening Neural Network models against adversarial attacks. Adaptive-Gravity is an iterative method. The corresponding latent spaces related to each head and tail layer are pushed further apart and concentrated toward their corresponding centroid at each iteration. To find the direction for moving the latent spaces apart, we introduced the anti-gravity force. We stipulate a new location for the centroids of each set of latent class features in the tail and head layers' latent spaces using the anti-gravity force. Subsequently, we deploy a training procedure to adjust the underlying model's parameters to achieve two objectives 1) improving the accuracy of the underlying model 2) concentrating the latent spaces of the tail and head layers to their projected centroids. We evaluated our method against diverse attack scenarios on CIFAR10 and MNIST datasets. Our experimental results indicate that Adaptive-Gravity is an extremely effective and resilient solution for resisting the existing adversarial attacks.  

\section{Acknowledgment}
This work was supported by National Science Foundation (NSF) under NSF award number 1718538.




\def\bibfont{\tiny}

\bibliographystyle{IEEEtran}

\bibliography {Gravity}

\end{document}